\pdfoutput=1

\documentclass[11pt]{article}

\usepackage{acl}

\usepackage{times}
\usepackage{latexsym}

\usepackage[T1]{fontenc}

\usepackage[utf8]{inputenc}

\usepackage{microtype}

\usepackage{inconsolata}

\usepackage{amsmath}
\usepackage{colortbl}
\usepackage{booktabs} %
\usepackage{multirow}
\usepackage{cite}
\usepackage{graphicx}
\usepackage{xurl}
\usepackage{subcaption}

\title{An Empirical Study of Speech Language Models for Prompt-Conditioned Speech Synthesis}

\author{Yifan Peng$^1$\thanks{\quad Work done during an internship at Meta AI.} , Ilia Kulikov$^2$, Yilin Yang$^2$, Sravya Popuri$^2$, Hui Lu$^{3*}$, \\{\bf Changhan Wang$^2$, Hongyu Gong$^2$} \\
$^1$ Carnegie Mellon University\quad
$^2$ Meta AI \quad
$^3$ The Chinese University of Hong Kong\\
}

\begin{document}
\maketitle
\begin{abstract}
Speech language models (LMs) are promising for high-quality speech synthesis through in-context learning. A typical speech LM takes discrete semantic units as content and a short utterance as prompt, and synthesizes speech which preserves the content's semantics but mimics the prompt's style. However, there is no systematic understanding on how the synthesized audio is controlled by the prompt and content. In this work, we conduct an empirical study of the widely used autoregressive (AR) and non-autoregressive (NAR) speech LMs and provide insights into the prompt design and content semantic units. Our analysis reveals that heterogeneous and nonstationary prompts hurt the audio quality in contrast to the previous finding that longer prompts always lead to better synthesis. Moreover, we find that the speaker style of the synthesized audio is also affected by the content in addition to the prompt. We further show that semantic units carry rich acoustic information such as pitch, tempo, volume and speech emphasis, which might be leaked from the content to the synthesized audio.
\end{abstract}

\section{Introduction}
\label{sec:intro}

Language models (LMs) have showcased strong in-context learning capabilities in natural language processing~\citep{brown2020language, chowdhery2022palm, touvron2023llama}.
Recent advances in audio quantization~\citep{soundstream, encodec} have opened an opportunity to utilize autoregressive (AR) LMs to generate high-quality natural speech by modeling the distribution over discrete speech units~\citep{audiolm, valle, vallex}. Another line of work directly models the distribution over continuous features such as Mel spectrograms or quantized features with non-autoregressive (NAR) models~\citep{voicebox, natural-speech-2}.
These speech LMs, trained on large amounts of speech data, demonstrate state-of-the-art (SOTA) performance in zero-shot conditional speech synthesis tasks, where the desired content is represented as a sequence of discrete units and the desired style is provided by a speech prompt of a few seconds.

Despite numerous studies on model architectures, training methods and downstream applications, there is no systematic understanding of how the prompt and content affect the synthesized speech in vocal style, emotion and prosody. It is unclear which attributes can be manipulated through prompts. For example, can we adapt the speech rate of the content to match that of the prompt by using a speech LM? Does it work for both AR and NAR LMs? Should we deduplicate content units to remove the original duration? Addressing these questions provides insights into the true capabilities of speech LMs, consequently offering valuable guidance for enhancing their performance.

This work aims to address the following questions for both AR and NAR speech LMs through quantitative analysis. We will publicly release the code for empirical evaluation. 

Prior studies show that longer prompts yield better speech style transfer results~\citep{valle, natural-speech-2, voicebox}. However, this implicitly assumes that the prompt always has consistent vocal style and speaker emotion regardless of its length. In practice,
longer prompts can become heterogeneous or nonstationary, which might adversely affect the synthesized speech. We address the following two questions in Section~\ref{sec:exp_prompt}:

\noindent \textbf{Q1.1}: How does a \textbf{heterogeneous} prompt containing multiple vocal styles from different speakers affect the generated speech?

\noindent \textbf{Q1.2}: How does a \textbf{nonstationary} prompt containing mixed emotions affect the generated speech?

Recent studies~\citep{audiolm, soundstorm, make-a-voice, polyvoice} represent the content of an utterance with semantic units from HuBERT~\citep{hubert} or w2v-BERT~\citep{W2v-bert}, assuming these units mostly contain semantic information only. But HuBERT units do contain other information~\citep{lin2023utility}.

\noindent\textbf{Q2.1}: Will other information in the content units be leaked to the synthesized speech? See Section~\ref{sec:exp_content}.

\noindent\textbf{Q2.2}: How do prompt and content control acoustic features of synthesized speech, like pitch, speech rate, volume and emphasis? See Section~\ref{sec:exp_prosody}.

\section{Related Work}

\begingroup
\setlength{\tabcolsep}{2pt}

\begin{table*}[t]
    \centering
    \resizebox {\linewidth} {!} {
    \begin{tabular}{cccccc}
    \toprule
    Name & Speech representation & Model type & Initialization & Supported tasks\\
    \midrule
    GSLM~\citep{lakhotia-etal-2021-generative} & SSL units & AR LM & - & Speech continuation \\
    pGSLM~\citep{kharitonov-etal-2022-text} & SSL units, F0, duration & AR LM & - & Speech continuation\\
    AudioLM~\citep{audiolm} & SSL units, codec units & AR LM & - & Speech continuation\\
    TWIST~\citep{twist} & SSL units & AR LM & Text LM & Speech continuation \\
    VALL-E~\citep{valle} & Codec units & AR LM, NAR LM & - & TTS \\
    VALL-E X~\citep{vallex} & Codec units & AR LM, NAR LM & - & TTS \\
    VioLA~\citep{viola} & Codec units & AR LM, NAR LM & - & ASR, MT, ST, TTS, S2ST\\
    MusicGen~\citep{music-gen} & Codec units & AR LM & - & Music generation\\
    AudioPaLM~\citep{audiopalm} & SSL/ASR units, codec units & AR LM, NAR LM & Text LM & ASR, MT, ST, TTS, S2ST\\
    SpeechX~\citep{speechx} & Codec units & AR LM, NAR LM & VALL-E & \begin{tabular}[c]{@{}c@{}}TTS, denoising, speech removal,\\ target speaker extraction, speech editing\end{tabular} \\
    VoxtLM~\citep{voxtlm} & SSL units & AR LM & Text LM & ASR, TTS, speech and text continuation \\
    \midrule
    SoundStorm~\citep{soundstorm} & SSL units, codec units & NAR LM & - & Speech continuation\\
    NaturalSpeech 2~\citep{natural-speech-2} & Continuous features & NAR diffusion & - & TTS\\
    Voicebox~\citep{voicebox} & Continuous features & NAR normalizing flow & - & TTS, noise removal, content editing, style conversion\\
    \bottomrule
    \end{tabular}
}
\vskip -0.1in
\caption{Summary of recent studies about speech LMs. More discussions are presented in \autoref{app:literature-review}.}
\label{tab:speechlm-papers}
\end{table*}

\endgroup

\label{subsec:overview-speech-lm}

\autoref{tab:speechlm-papers} compares recent speech LMs from various aspects. In this section, we provide a brief summary of the key aspects. More details are in \autoref{app:literature-review}.

\noindent\textbf{Speech units.} There has been a lot of progress on discrete speech representations including semantic units which mainly capture semantic information~\citep{hubert} and acoustic units which contain rich acoustic features~\citep{encodec}.

\noindent\textbf{Initialization.} Speech LMs can be initialized with pre-trained text LMs %
to improve performance~\citep{twist, audiopalm}.

\begin{figure*}[t]
     \centering
     \hfill
     \begin{subfigure}[t]{0.25\linewidth}
         \centering
         \includegraphics[scale=0.43]{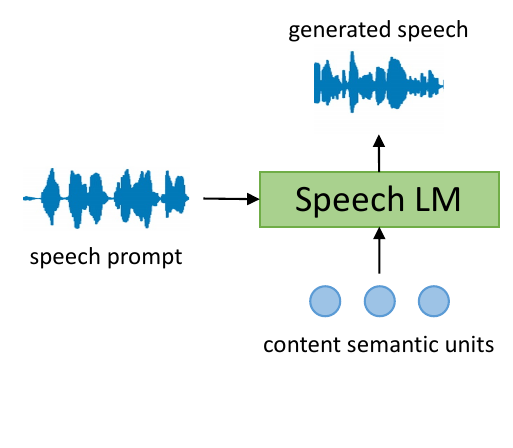}
         \vskip -0.08in
         \caption{Conditional speech synthesis}
         \label{fig:task}
     \end{subfigure}
     \hfill
     \begin{subfigure}[t]{0.48\linewidth}
         \centering
         \includegraphics[scale=0.43]{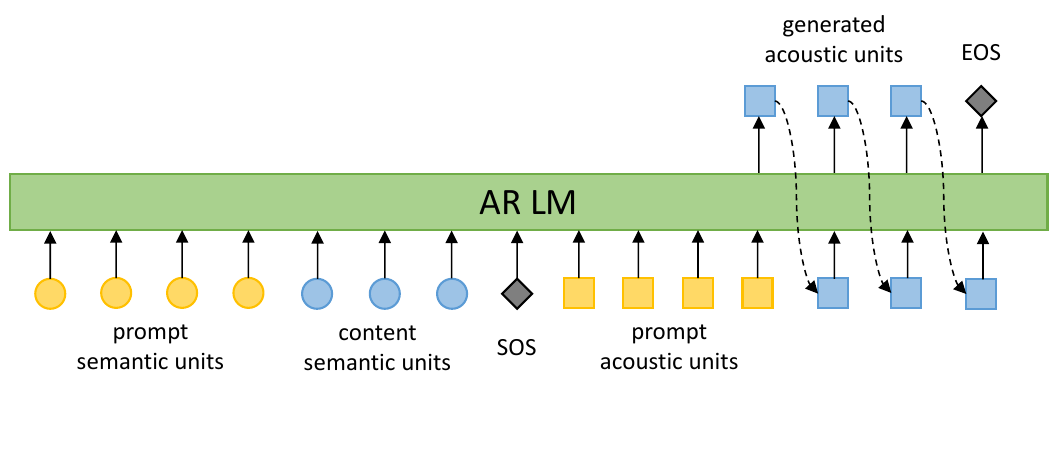}
         \vskip -0.08in
         \caption{Inference of AR LM}
         \label{fig:ar-lm}
     \end{subfigure}
     \hfill
     \begin{subfigure}[t]{0.25\linewidth}
         \centering
         \includegraphics[scale=0.43]{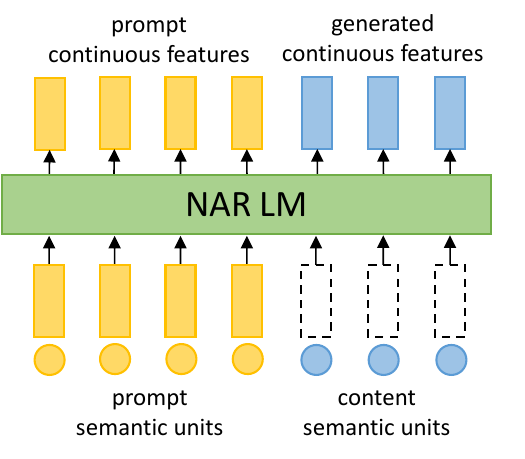}
         \vskip -0.08in
         \caption{Inference of NAR LM}
         \label{fig:nar-lm}
     \end{subfigure}
     \hfill
     \vskip -0.1in
    \caption{Overview of the primary task of speech LMs and inference procedures of AR and NAR LMs.}
    \label{fig:overview}
\end{figure*}

\noindent\textbf{AR vs NAR LMs.}  \autoref{fig:ar-lm} shows the inference of AR LMs. We study the VALL-E style model~\citep{valle}
and consider two variants of AR LMs: one with duplicate semantic units and the other with deduplicated semantic units.
\autoref{fig:nar-lm} illustrates NAR LMs. We analyze Voicebox~\citep{voicebox} as it achieves SOTA performance in various conditional speech synthesis tasks.

\section{Experiments}

We analyze both AR and NAR LMs on conditional speech synthesis (see \autoref{fig:task}), which is one of the primary tasks of speech LMs. Specifically, a speech LM takes as input a pair of prompt and content utterances, and synthesizes a new utterance that mimics the style of the prompt but preserves the semantic meaning of the content. This task is also referred to as voice conversion or style transfer. Following recent studies~\citep{audiolm, soundstorm, polyvoice}, we represent content with HuBERT units~\citep{hubert}. %

\subsection{Experimental setups}

\noindent\textbf{Training data}. We use 60k-hour English speech as training data of Speech LMs.

\noindent\textbf{Evaluation data}. 
We create various analysis data using emotional English speech with transcriptions. The speech is collected from multiple emotions including neutral, amused, sleepy, angry and disgust.
We also prepare some samples for emphasis analysis.
We will discuss more about data preparation in corresponding sections.

\noindent\textbf{Evaluation of speaker style similarity}. We employ a WavLM~\citep{wavlm} based speaker style encoder to generate the speaker style embedding for a given utterance. We then calculate the cosine similarity between a pair of speaker style embeddings as the speaker style similarity. This is a standard evaluation metric used in prior work~\citep{valle, voicebox}.

\noindent\textbf{Speech tokenizers}. Semantic units are derived from HuBERT~\citep{hubert} and acoustic units are extracted from EnCodec~\citep{encodec} with 8 codebooks. Both are trained on VoxPopuli~\citep{wang-etal-2021-voxpopuli} with 50Hz unit rate.

\noindent\textbf{Speech LMs}. We use fairseq~\citep{ott-etal-2019-fairseq} for implementation. The training details are provided in Appendix~\ref{app:lm-training}.

(1) \textbf{AR LM with duplicate semantic units.} We follow VALL-E~\citep{valle} but replace its text condition with semantic units. The AR LM is a 24-layer Transformer decoder with embedding dimension 1024 and feed-forward dimension 4096.\footnote{The AR LM predicts 1st EnCodec stream conditioned on semantic units. Similar to VALL-E, a secondary NAR LM of the same size is trained to predict the remaining streams.}
 
(2) \textbf{AR LM with deduplicated semantic units.} The model is the same as (1), but semantic units are deduplicated to remove some duration information.

(3) \textbf{NAR LM.} We follow Voicebox~\citep{voicebox} but replace the text condition with semantic units. It has 24 Transformer layers with embedding dimension $1024$ and feed-forward dimension $4096$.

\begin{table}[tbp!]
\centering
\resizebox {\linewidth} {!} {
\begin{tabular}{ccccc}
\toprule
\textbf{Prompt used for synthesis} & \multicolumn{2}{c}{\textbf{P1}} & \multicolumn{2}{c}{\textbf{P1+P2}} \\ 
\textbf{Reference audio} & P1 & P2 & P1 & P2  \\ \midrule
\textbf{AR w/ dup units} & 0.332 & 0.036 & 0.080 & 0.135 \\
\textbf{AR w/ dedup units} & 0.345 & 0.054 & 0.098 & 0.163 \\
\textbf{NAR} & 0.455 & 0.062 & 0.105 & 0.285 \\ 
\bottomrule
\end{tabular}
}
\vskip -0.1in
\caption{Speaker style similarity ($\uparrow$) between the synthesized audio and each prompt audio for \textbf{heterogeneous} prompts. P1 and P2 are prompts from different speaker styles. P1+P2 is the concatenation of P1 and P2. Results are averaged over 400 evaluation samples.}
\label{tab:prompt_het_sim}
\end{table}

\subsection{Effect of heterogeneous and nonstationary prompts}
\label{sec:exp_prompt}

Previous studies~\citep{valle, natural-speech-2, voicebox} show that a longer prompt consistently improves synthesis. In practice, a longer prompt is more likely to become inconsistent in vocal style and emotion. Hence, we consider two properties of the prompt and examine their impacts on conditional speech synthesis: (1) \textbf{heterogeneity}, where a prompt contains multiple styles of different speakers, and (2) \textbf{nonstationarity}, where a prompt is from the same speaker style but mixed by different emotions.

\noindent\textbf{Heterogeneity}.
We concatenate audios of two speaker styles in the emotional data to form heterogeneous prompts. As a controlled study, the semantic contents (transcriptions) of both prompt and content audios are identical, and their emotions are also the same (i.e., neutral). The evaluation set has $400$ samples. To evaluate synthesized audios, we report their speaker similarity w.r.t. the two prompt audios respectively. For comparison, we also synthesize audios with a single prompt and the same content audio used in the multi-prompt experiments. This can reflect the difference between single-speaker-style and multi-speaker-style prompts. As shown in \autoref{tab:prompt_het_sim}, multi-speaker-style prompt hurts the speaker style similarity. More discussions are in Appendix~\ref{app:effect-heterogeneous}.

\begin{table}[tbp!]
\centering
\resizebox {\linewidth} {!} {
\begin{tabular}{ccccc}
\toprule
\textbf{Prompt used for synthesis} & \multicolumn{2}{c}{\textbf{P1}} & \multicolumn{2}{c}{\textbf{P1+P2}} \\ 
\textbf{Reference audio} & P1 & P2 & P1 & P2 \\ \midrule
\textbf{AR w/ dup units} & 0.257 & 0.073 & 0.133 & 0.156 \\
\textbf{AR w/ dedup units} & 0.256 & 0.073 & 0.140 & 0.165 \\
\textbf{NAR} & 0.392 & 0.160 & 0.222 & 0.336  \\ 
\bottomrule
\end{tabular}
}
\vskip -0.1in
\caption{Speaker style similarity ($\uparrow$) between the synthesized audio and prompt audio for \textbf{nonstationary} prompts. P1 and P2 are in the same style but different emotions. P1+P2 is the concatenation of P1 and P2. Results are averaged over 400 samples.}
\label{tab:prompt_ns_spkr}
\end{table}

\noindent\textbf{Nonstationarity}.
We concatenate audios from the same speaker style (i.e., vocal style) but with different emotions (e.g., amused and sleepy) to form nonstationary prompts, resulting in totally $400$ evaluation samples. \autoref{tab:prompt_ns_spkr} shows that mixed-emotion prompt also hurts the speaker style similarity. More discussions are in Appendix~\ref{app:effect-nonstationary}.

\begingroup
\setlength{\tabcolsep}{3pt}

\begin{table}[tbp!]
\centering
\resizebox {\linewidth} {!} {
\begin{tabular}{ccccccc}
\toprule
\textbf{Style of content audio} & \multicolumn{2}{c}{\textbf{F2}} & \multicolumn{2}{c}{\textbf{M1}} & \multicolumn{2}{c}{\textbf{M2}} \\ 
\textbf{Reference audio} & Prompt & Content & Prompt & Content & Prompt & Content \\ \midrule
AR w/ dup units & 0.535 & 0.179 & 0.488 & 0.125 & 0.489 & 0.046 \\
AR w/ dedup units & 0.536 & 0.151 & 0.474 & 0.118 & 0.489 & 0.038 \\
NAR & 0.584 & 0.222 & 0.534 & 0.148 & 0.529 & 0.106 \\ 
\bottomrule
\end{tabular}
}
\vskip -0.1in
\caption{Speaker style similarity ($\uparrow$) between the synthesized audio and the prompt or content audio. The prompt is fixed as the female speaker style F1, while the content style is changed among F2 (female), M1 (male), and M2 (male). We can observe that the content speaker style also affects the synthesized style.}
\label{tab:content_spkr}
\end{table}

\endgroup

\subsection{Effect of content audio's speaker styles}
\label{sec:exp_content}

Although existing studies use prompt audio to control synthesized vocal style~\citep{soundstorm, make-a-voice}, it remains unexplored how much content audio affects the vocal style in speech synthesis.
To investigate this, we prepare an evaluation set of $200$ samples using emotional speech data, where we fix the speaker style of prompt audios as a female speaker, F1, but change the speaker style of content audios among another female F2 and two male speaker styles M1 and M2. The synthesized audios are compared with the prompt and content audios respectively in terms of speaker style similarity. In \autoref{tab:content_spkr}, it is found that the change of content speaker style results in different voices, indicating that semantic units like HuBERT carry more acoustic information than expected. Appendix~\ref{app:effect-content-spkr} includes more discussions.

\begingroup
\setlength{\tabcolsep}{3pt}

\begin{table}[tbp!]
\centering
\resizebox {\linewidth} {!} {

\begin{tabular}{ccccccc}
\toprule
\textbf{Changed prosody feature} & \multicolumn{2}{c}{\textbf{Pitch}} & \multicolumn{2}{c}{\textbf{Tempo}} & \multicolumn{2}{c}{\textbf{Volume}} \\
Changed audio & Prompt & Content & Prompt & Content & Prompt & Content \\
\midrule
AR w/ dup units & 0.293 &  -0.037 & 0.054 & 0.822 & 0.987 & 0.025 \\
AR w/ dedup units & 0.136 & 0.001 & -0.023 & 0.388 & 0.982 & 0.053 \\
NAR & 0.476 & 0.135 & 0.000 & 0.999 & 0.997 & 0.217 \\

\bottomrule
\end{tabular}
}
\vskip -0.1in
\caption{Pearson correlation of prosody changes between the synthesized audio and either the prompt or content audio. Each experiment only changes one feature of either prompt or content.
}
\label{tab:prosody_corr}
\end{table}

\endgroup

\subsection{Analysis of prosody information}
\label{sec:exp_prosody}
Our analyses so far focus on voice and style transfer based on a coarse-grained metric, speaker style similarity. Now we look deeper into fine-grained acoustic features including pitch, speech rate (tempo), loudness (volume) and emphasis. 
A set of $200$ audio samples is selected from the emotional data for such prosody analysis.
We first use the same audio as prompt and content to synthesize a set of audios with speech LMs, which serves as the reference set since no manipulation is performed. Then, we manually manipulate the acoustic characteristics of either prompt or content audios. The Pearson correlation of acoustic changes between prompt/content and generated audios is reported in \autoref{tab:prosody_corr}. Please refer to Appendix~\ref{app:analysis-prosody} for more discussions.

\textbf{Pitch}. We use torchaudio pitch extractor\footnote{\url{https://pytorch.org/audio/2.0.1/tutorials/audio_feature_extractions_tutorial.html\#pitch}} to extract pitch.
We observe that AR LMs capture pitch information mostly from its prompt. NAR LMs are affected by the pitch of both prompt and content, which also indicates that content semantic units carry some pitch information.

\textbf{Speech rate}. We measure the number of syllables\footnote{Python syllable estimator: \url{https://pypi.org/project/syllables/}.} spoken per second.
We find that the speech rate (tempo) is mainly determined by content units for both AR and NAR LMs.
The AR LM with deduplicated units has a lower correlation, suggesting that it can generate more flexible or diverse speech rates.
However, \textbf{the speech rate cannot be controlled by prompts in current speech LMs}.

\textbf{Loudness}. We use pyloudnorm~\citep{steinmetz2021pyloudnorm} to measure loudness.
We observe that the volume of synthesized audio is mainly determined by prompt, while the NAR LM also transfers loudness of the content to its synthesized audio.

Finally, we analyze how \textbf{speech emphasis} is affected by the prompt or content audio.
To examine word emphasis, we take $50$ audio sample pairs.
Each pair of prompt and content audios has the same semantic meaning and is spoken in the same speaker style. The difference is that some words are emphasized in the content audio while the prompt does not have any speech emphasis. We aim to study whether the emphasis is embedded in the content semantic units and further transferred to synthesized audio. 

Two annotators are asked to check whether the synthesized speech has the same emphasis as the content audio and go through annotations together to resolve disagreements. 
The percentages of synthesize audios preserving content emphasis are $96\%$, $80\%$ and $98\%$ for AR LM w/ dup units, AR w/ dedup units and NAR LM, respectively. It indicates that content semantic units do carry emphasis information which is further leaked to synthesized audios. \textbf{Current speech LMs cannot directly control speech emphasis through prompts.}

\section{Conclusion}
We conduct an empirical study of AR and NAR speech LMs for speech synthesis conditioned on prompt and semantic units. 
We reveal that heterogeneous and nonstationary prompts can hurt vocal style transfer.
We also find that content audio style affects the synthesized vocal style through semantic units.
In particular, we show that semantic units of content audio carry rich information like pitch, tempo, volume and speech emphasis, which might be leaked to the synthesized audio. 
These findings indicate that contemporary speech LMs using semantic units cannot achieve
zero-shot style transfer or controllable speech synthesis solely through prompts. 
Future research can explore more disentangled discrete speech representations and better modeling algorithms.

\section{Limitations}

\textbf{Limitations}. In this work, we designed a set of tasks to benchmark speech LMs in the task of conditional speech synthesis. The evaluation may not be comprehensive, and other metrics such as speech naturalness could be incorporated into future study.

\textbf{Ethical considerations.} While we have documented various evaluation deployed in our work, here are some additional points to highlight. While high-quality speech synthesis could improve real-world applications and facilitate communication, such access could also make groups with lower levels of digital literacy more vulnerable to misinformation. An example of unintended use is that bad actors misappropriate our work for online scams.

\bibliography{anthology,custom}

\begin{thebibliography}{40}
\expandafter\ifx\csname natexlab\endcsname\relax\def\natexlab#1{#1}\fi

\bibitem[{Borsos et~al.(2023{\natexlab{a}})Borsos, Marinier, Vincent, Kharitonov, Pietquin, Sharifi, Roblek, Teboul, Grangier, Tagliasacchi, and Zeghidour}]{audiolm}
Zal{\'{a}}n Borsos, Rapha{\"{e}}l Marinier, Damien Vincent, Eugene Kharitonov, Olivier Pietquin, Matthew Sharifi, Dominik Roblek, Olivier Teboul, David Grangier, Marco Tagliasacchi, and Neil Zeghidour. 2023{\natexlab{a}}.
\newblock \href {https://doi.org/10.1109/TASLP.2023.3288409} {Audiolm: {A} language modeling approach to audio generation}.
\newblock \emph{{IEEE} {ACM} Trans. Audio Speech Lang. Process.}, 31:2523--2533.

\bibitem[{Borsos et~al.(2023{\natexlab{b}})Borsos, Sharifi, Vincent, Kharitonov, Zeghidour, and Tagliasacchi}]{soundstorm}
Zal{\'{a}}n Borsos, Matthew Sharifi, Damien Vincent, Eugene Kharitonov, Neil Zeghidour, and Marco Tagliasacchi. 2023{\natexlab{b}}.
\newblock \href {https://doi.org/10.48550/arXiv.2305.09636} {Soundstorm: Efficient parallel audio generation}.
\newblock \emph{CoRR}, abs/2305.09636.

\bibitem[{Brown et~al.(2020{\natexlab{a}})Brown, Mann, Ryder, Subbiah, Kaplan, Dhariwal, Neelakantan, Shyam, Sastry, Askell, Agarwal, Herbert-Voss, Krueger, Henighan, Child, Ramesh, Ziegler, Wu, Winter, Hesse, Chen, Sigler, Litwin, Gray, Chess, Clark, Berner, McCandlish, Radford, Sutskever, and Amodei}]{brown2020language}
Tom Brown, Benjamin Mann, Nick Ryder, Melanie Subbiah, Jared~D Kaplan, Prafulla Dhariwal, Arvind Neelakantan, Pranav Shyam, Girish Sastry, Amanda Askell, Sandhini Agarwal, Ariel Herbert-Voss, Gretchen Krueger, Tom Henighan, Rewon Child, Aditya Ramesh, Daniel Ziegler, Jeffrey Wu, Clemens Winter, Chris Hesse, Mark Chen, Eric Sigler, Mateusz Litwin, Scott Gray, Benjamin Chess, Jack Clark, Christopher Berner, Sam McCandlish, Alec Radford, Ilya Sutskever, and Dario Amodei. 2020{\natexlab{a}}.
\newblock \href {https://proceedings.neurips.cc/paper_files/paper/2020/file/1457c0d6bfcb4967418bfb8ac142f64a-Paper.pdf} {Language models are few-shot learners}.
\newblock In \emph{Advances in Neural Information Processing Systems}, volume~33, pages 1877--1901. Curran Associates, Inc.

\bibitem[{Brown et~al.(2020{\natexlab{b}})Brown, Mann, Ryder, Subbiah, Kaplan, Dhariwal, Neelakantan, Shyam, Sastry, Askell, Agarwal, Herbert{-}Voss, Krueger, Henighan, Child, Ramesh, Ziegler, Wu, Winter, Hesse, Chen, Sigler, Litwin, Gray, Chess, Clark, Berner, McCandlish, Radford, Sutskever, and Amodei}]{DBLP:conf/nips/BrownMRSKDNSSAA20}
Tom~B. Brown, Benjamin Mann, Nick Ryder, Melanie Subbiah, Jared Kaplan, Prafulla Dhariwal, Arvind Neelakantan, Pranav Shyam, Girish Sastry, Amanda Askell, Sandhini Agarwal, Ariel Herbert{-}Voss, Gretchen Krueger, Tom Henighan, Rewon Child, Aditya Ramesh, Daniel~M. Ziegler, Jeffrey Wu, Clemens Winter, Christopher Hesse, Mark Chen, Eric Sigler, Mateusz Litwin, Scott Gray, Benjamin Chess, Jack Clark, Christopher Berner, Sam McCandlish, Alec Radford, Ilya Sutskever, and Dario Amodei. 2020{\natexlab{b}}.
\newblock \href {https://proceedings.neurips.cc/paper/2020/hash/1457c0d6bfcb4967418bfb8ac142f64a-Abstract.html} {Language models are few-shot learners}.
\newblock In \emph{Advances in Neural Information Processing Systems 33: Annual Conference on Neural Information Processing Systems 2020, NeurIPS 2020, December 6-12, 2020, virtual}.

\bibitem[{Chen et~al.(2022)Chen, Wang, Chen, Wu, Liu, Chen, Li, Kanda, Yoshioka, Xiao, Wu, Zhou, Ren, Qian, Qian, Wu, Zeng, Yu, and Wei}]{wavlm}
Sanyuan Chen, Chengyi Wang, Zhengyang Chen, Yu~Wu, Shujie Liu, Zhuo Chen, Jinyu Li, Naoyuki Kanda, Takuya Yoshioka, Xiong Xiao, Jian Wu, Long Zhou, Shuo Ren, Yanmin Qian, Yao Qian, Jian Wu, Michael Zeng, Xiangzhan Yu, and Furu Wei. 2022.
\newblock \href {https://doi.org/10.1109/JSTSP.2022.3188113} {Wavlm: Large-scale self-supervised pre-training for full stack speech processing}.
\newblock \emph{{IEEE} J. Sel. Top. Signal Process.}, 16(6):1505--1518.

\bibitem[{Chowdhery et~al.(2022)Chowdhery, Narang, Devlin, Bosma, Mishra, Roberts, Barham, Chung, Sutton, Gehrmann, Schuh, Shi, Tsvyashchenko, Maynez, Rao, Barnes, Tay, Shazeer, Prabhakaran, Reif, Du, Hutchinson, Pope, Bradbury, Austin, Isard, Gur{-}Ari, Yin, Duke, Levskaya, Ghemawat, Dev, Michalewski, Garcia, Misra, Robinson, Fedus, Zhou, Ippolito, Luan, Lim, Zoph, Spiridonov, Sepassi, Dohan, Agrawal, Omernick, Dai, Pillai, Pellat, Lewkowycz, Moreira, Child, Polozov, Lee, Zhou, Wang, Saeta, Diaz, Firat, Catasta, Wei, Meier{-}Hellstern, Eck, Dean, Petrov, and Fiedel}]{chowdhery2022palm}
Aakanksha Chowdhery, Sharan Narang, Jacob Devlin, Maarten Bosma, Gaurav Mishra, Adam Roberts, Paul Barham, Hyung~Won Chung, Charles Sutton, Sebastian Gehrmann, Parker Schuh, Kensen Shi, Sasha Tsvyashchenko, Joshua Maynez, Abhishek Rao, Parker Barnes, Yi~Tay, Noam Shazeer, Vinodkumar Prabhakaran, Emily Reif, Nan Du, Ben Hutchinson, Reiner Pope, James Bradbury, Jacob Austin, Michael Isard, Guy Gur{-}Ari, Pengcheng Yin, Toju Duke, Anselm Levskaya, Sanjay Ghemawat, Sunipa Dev, Henryk Michalewski, Xavier Garcia, Vedant Misra, Kevin Robinson, Liam Fedus, Denny Zhou, Daphne Ippolito, David Luan, Hyeontaek Lim, Barret Zoph, Alexander Spiridonov, Ryan Sepassi, David Dohan, Shivani Agrawal, Mark Omernick, Andrew~M. Dai, Thanumalayan~Sankaranarayana Pillai, Marie Pellat, Aitor Lewkowycz, Erica Moreira, Rewon Child, Oleksandr Polozov, Katherine Lee, Zongwei Zhou, Xuezhi Wang, Brennan Saeta, Mark Diaz, Orhan Firat, Michele Catasta, Jason Wei, Kathy Meier{-}Hellstern, Douglas Eck, Jeff Dean, Slav Petrov, and Noah Fiedel.
  2022.
\newblock \href {https://doi.org/10.48550/arXiv.2204.02311} {Palm: Scaling language modeling with pathways}.
\newblock \emph{CoRR}, abs/2204.02311.

\bibitem[{Chung et~al.(2021)Chung, Zhang, Han, Chiu, Qin, Pang, and Wu}]{W2v-bert}
Yu{-}An Chung, Yu~Zhang, Wei Han, Chung{-}Cheng Chiu, James Qin, Ruoming Pang, and Yonghui Wu. 2021.
\newblock \href {https://doi.org/10.1109/ASRU51503.2021.9688253} {w2v-bert: Combining contrastive learning and masked language modeling for self-supervised speech pre-training}.
\newblock In \emph{{IEEE} Automatic Speech Recognition and Understanding Workshop, {ASRU} 2021, Cartagena, Colombia, December 13-17, 2021}, pages 244--250. {IEEE}.

\bibitem[{Communication et~al.(2023)Communication, Barrault, Chung, Meglioli, Dale, Dong, Duquenne, Elsahar, Gong, Heffernan, Hoffman, Klaiber, Li, Licht, Maillard, Rakotoarison, Sadagopan, Wenzek, Ye, Akula, Chen, Hachem, Ellis, Gonzalez, Haaheim, Hansanti, Howes, Huang, Hwang, Inaguma, Jain, Kalbassi, Kallet, Kulikov, Lam, Li, Ma, Mavlyutov, Peloquin, Ramadan, Ramakrishnan, Sun, Tran, Tran, Tufanov, Vogeti, Wood, Yang, Yu, Andrews, Balioglu, Costa{-}juss{\`{a}}, Celebi, Elbayad, Gao, Guzm{\'{a}}n, Kao, Lee, Mourachko, Pino, Popuri, Ropers, Saleem, Schwenk, Tomasello, Wang, Wang, and Wang}]{seamlessm4t}
Seamless Communication, Lo{\"{\i}}c Barrault, Yu{-}An Chung, Mariano~Cora Meglioli, David Dale, Ning Dong, Paul{-}Ambroise Duquenne, Hady Elsahar, Hongyu Gong, Kevin Heffernan, John Hoffman, Christopher Klaiber, Pengwei Li, Daniel Licht, Jean Maillard, Alice Rakotoarison, Kaushik~Ram Sadagopan, Guillaume Wenzek, Ethan Ye, Bapi Akula, Peng{-}Jen Chen, Naji~El Hachem, Brian Ellis, Gabriel~Mejia Gonzalez, Justin Haaheim, Prangthip Hansanti, Russ Howes, Bernie Huang, Min{-}Jae Hwang, Hirofumi Inaguma, Somya Jain, Elahe Kalbassi, Amanda Kallet, Ilia Kulikov, Janice Lam, Daniel Li, Xutai Ma, Ruslan Mavlyutov, Benjamin Peloquin, Mohamed Ramadan, Abinesh Ramakrishnan, Anna~Y. Sun, Kevin Tran, Tuan Tran, Igor Tufanov, Vish Vogeti, Carleigh Wood, Yilin Yang, Bokai Yu, Pierre Andrews, Can Balioglu, Marta~R. Costa{-}juss{\`{a}}, Onur Celebi, Maha Elbayad, Cynthia Gao, Francisco Guzm{\'{a}}n, Justine Kao, Ann Lee, Alexandre Mourachko, Juan Pino, Sravya Popuri, Christophe Ropers, Safiyyah Saleem, Holger Schwenk, Paden
  Tomasello, Changhan Wang, Jeff Wang, and Skyler Wang. 2023.
\newblock \href {https://doi.org/10.48550/ARXIV.2308.11596} {Seamlessm4t-massively multilingual {\&} multimodal machine translation}.
\newblock \emph{CoRR}, abs/2308.11596.

\bibitem[{Copet et~al.(2023)Copet, Kreuk, Gat, Remez, Kant, Synnaeve, Adi, and D{\'{e}}fossez}]{music-gen}
Jade Copet, Felix Kreuk, Itai Gat, Tal Remez, David Kant, Gabriel Synnaeve, Yossi Adi, and Alexandre D{\'{e}}fossez. 2023.
\newblock \href {https://doi.org/10.48550/arXiv.2306.05284} {Simple and controllable music generation}.
\newblock \emph{CoRR}, abs/2306.05284.

\bibitem[{D{\'{e}}fossez et~al.(2022)D{\'{e}}fossez, Copet, Synnaeve, and Adi}]{encodec}
Alexandre D{\'{e}}fossez, Jade Copet, Gabriel Synnaeve, and Yossi Adi. 2022.
\newblock \href {https://doi.org/10.48550/arXiv.2210.13438} {High fidelity neural audio compression}.
\newblock \emph{CoRR}, abs/2210.13438.

\bibitem[{Dong et~al.(2023)Dong, Huang, Tian, Xu, Zhao, Wang, Cheng, Ko, Tian, Li, Yue, Bai, Chen, Lu, Ma, Wang, Wang, and Wang}]{polyvoice}
Qianqian Dong, Zhiying Huang, Qiao Tian, Chen Xu, Yunlong Zhao, Kexin Wang, Xuxin Cheng, Tom Ko, Qiao Tian, Tang Li, Fengpeng Yue, Ye~Bai, Xi~Chen, Lu~Lu, Zejun Ma, Yuping Wang, Mingxuan Wang, and Yuxuan Wang. 2023.
\newblock \href {https://doi.org/10.48550/arXiv.2306.02982} {Polyvoice: Language models for speech to speech translation}.
\newblock \emph{CoRR}, abs/2306.02982.

\bibitem[{Hassid et~al.(2023)Hassid, Remez, Nguyen, Gat, Conneau, Kreuk, Copet, D{\'{e}}fossez, Synnaeve, Dupoux, Schwartz, and Adi}]{twist}
Michael Hassid, Tal Remez, Tu~Anh Nguyen, Itai Gat, Alexis Conneau, Felix Kreuk, Jade Copet, Alexandre D{\'{e}}fossez, Gabriel Synnaeve, Emmanuel Dupoux, Roy Schwartz, and Yossi Adi. 2023.
\newblock \href {https://doi.org/10.48550/arXiv.2305.13009} {Textually pretrained speech language models}.
\newblock \emph{CoRR}, abs/2305.13009.

\bibitem[{Hsu et~al.(2021)Hsu, Bolte, Tsai, Lakhotia, Salakhutdinov, and Mohamed}]{hubert}
Wei{-}Ning Hsu, Benjamin Bolte, Yao{-}Hung~Hubert Tsai, Kushal Lakhotia, Ruslan Salakhutdinov, and Abdelrahman Mohamed. 2021.
\newblock \href {https://doi.org/10.1109/TASLP.2021.3122291} {Hubert: Self-supervised speech representation learning by masked prediction of hidden units}.
\newblock \emph{{IEEE} {ACM} Trans. Audio Speech Lang. Process.}, 29:3451--3460.

\bibitem[{Huang et~al.(2023)Huang, Zhang, Wang, Yang, Liu, Ye, Jiang, Weng, Zhao, and Yu}]{make-a-voice}
Rongjie Huang, Chunlei Zhang, Yongqi Wang, Dongchao Yang, Luping Liu, Zhenhui Ye, Ziyue Jiang, Chao Weng, Zhou Zhao, and Dong Yu. 2023.
\newblock \href {https://doi.org/10.48550/arXiv.2305.19269} {Make-a-voice: Unified voice synthesis with discrete representation}.
\newblock \emph{CoRR}, abs/2305.19269.

\bibitem[{Kharitonov et~al.(2022)Kharitonov, Lee, Polyak, Adi, Copet, Lakhotia, Nguyen, Riviere, Mohamed, Dupoux, and Hsu}]{kharitonov-etal-2022-text}
Eugene Kharitonov, Ann Lee, Adam Polyak, Yossi Adi, Jade Copet, Kushal Lakhotia, Tu~Anh Nguyen, Morgane Riviere, Abdelrahman Mohamed, Emmanuel Dupoux, and Wei-Ning Hsu. 2022.
\newblock \href {https://doi.org/10.18653/v1/2022.acl-long.593} {Text-free prosody-aware generative spoken language modeling}.
\newblock In \emph{Proceedings of the 60th Annual Meeting of the Association for Computational Linguistics (Volume 1: Long Papers)}, pages 8666--8681, Dublin, Ireland. Association for Computational Linguistics.

\bibitem[{Kingma and Ba(2015)}]{adam}
Diederik~P. Kingma and Jimmy Ba. 2015.
\newblock \href {http://arxiv.org/abs/1412.6980} {Adam: {A} method for stochastic optimization}.
\newblock In \emph{3rd International Conference on Learning Representations, {ICLR} 2015, San Diego, CA, USA, May 7-9, 2015, Conference Track Proceedings}.

\bibitem[{Kong et~al.(2020)Kong, Kim, and Bae}]{hifi-gan}
Jungil Kong, Jaehyeon Kim, and Jaekyoung Bae. 2020.
\newblock \href {https://proceedings.neurips.cc/paper/2020/hash/c5d736809766d46260d816d8dbc9eb44-Abstract.html} {Hifi-gan: Generative adversarial networks for efficient and high fidelity speech synthesis}.
\newblock In \emph{Advances in Neural Information Processing Systems 33: Annual Conference on Neural Information Processing Systems 2020, NeurIPS 2020, December 6-12, 2020, virtual}.

\bibitem[{Lakhotia et~al.(2021)Lakhotia, Kharitonov, Hsu, Adi, Polyak, Bolte, Nguyen, Copet, Baevski, Mohamed, and Dupoux}]{lakhotia-etal-2021-generative}
Kushal Lakhotia, Eugene Kharitonov, Wei-Ning Hsu, Yossi Adi, Adam Polyak, Benjamin Bolte, Tu-Anh Nguyen, Jade Copet, Alexei Baevski, Abdelrahman Mohamed, and Emmanuel Dupoux. 2021.
\newblock \href {https://doi.org/10.1162/tacl_a_00430} {On generative spoken language modeling from raw audio}.
\newblock \emph{Transactions of the Association for Computational Linguistics}, 9:1336--1354.

\bibitem[{Le et~al.(2023)Le, Vyas, Shi, Karrer, Sari, Moritz, Williamson, Manohar, Adi, Mahadeokar, and Hsu}]{voicebox}
Matthew Le, Apoorv Vyas, Bowen Shi, Brian Karrer, Leda Sari, Rashel Moritz, Mary Williamson, Vimal Manohar, Yossi Adi, Jay Mahadeokar, and Wei{-}Ning Hsu. 2023.
\newblock \href {https://doi.org/10.48550/arXiv.2306.15687} {Voicebox: Text-guided multilingual universal speech generation at scale}.
\newblock \emph{CoRR}, abs/2306.15687.

\bibitem[{Lin et~al.(2022)Lin, Feng, Huang, Tseng, Lin, Li, Lee, and Ward}]{lin2023utility}
Guan{-}Ting Lin, Chi{-}Luen Feng, Wei{-}Ping Huang, Yuan Tseng, Tzu{-}Han Lin, Chen{-}An Li, Hung{-}yi Lee, and Nigel~G. Ward. 2022.
\newblock \href {https://doi.org/10.1109/SLT54892.2023.10023234} {On the utility of self-supervised models for prosody-related tasks}.
\newblock In \emph{{IEEE} Spoken Language Technology Workshop, {SLT} 2022, Doha, Qatar, January 9-12, 2023}, pages 1104--1111. {IEEE}.

\bibitem[{Maiti et~al.(2023{\natexlab{a}})Maiti, Peng, Choi, Jung, Chang, and Watanabe}]{voxtlm}
Soumi Maiti, Yifan Peng, Shukjae Choi, Jee{-}weon Jung, Xuankai Chang, and Shinji Watanabe. 2023{\natexlab{a}}.
\newblock \href {https://doi.org/10.48550/arXiv.2309.07937} {Voxtlm: unified decoder-only models for consolidating speech recognition/synthesis and speech/text continuation tasks}.
\newblock \emph{CoRR}, abs/2309.07937.

\bibitem[{Maiti et~al.(2023{\natexlab{b}})Maiti, Peng, Saeki, and Watanabe}]{Speechlmscore}
Soumi Maiti, Yifan Peng, Takaaki Saeki, and Shinji Watanabe. 2023{\natexlab{b}}.
\newblock \href {https://doi.org/10.1109/ICASSP49357.2023.10095710} {Speechlmscore: Evaluating speech generation using speech language model}.
\newblock In \emph{ICASSP 2023 - 2023 IEEE International Conference on Acoustics, Speech and Signal Processing (ICASSP)}, pages 1--5.

\bibitem[{OpenAI(2023)}]{gpt-4}
OpenAI. 2023.
\newblock \href {https://doi.org/10.48550/ARXIV.2303.08774} {{GPT-4} technical report}.
\newblock \emph{CoRR}, abs/2303.08774.

\bibitem[{Ott et~al.(2019)Ott, Edunov, Baevski, Fan, Gross, Ng, Grangier, and Auli}]{ott-etal-2019-fairseq}
Myle Ott, Sergey Edunov, Alexei Baevski, Angela Fan, Sam Gross, Nathan Ng, David Grangier, and Michael Auli. 2019.
\newblock \href {https://doi.org/10.18653/v1/N19-4009} {fairseq: A fast, extensible toolkit for sequence modeling}.
\newblock In \emph{Proceedings of the 2019 Conference of the North {A}merican Chapter of the Association for Computational Linguistics (Demonstrations)}, pages 48--53, Minneapolis, Minnesota. Association for Computational Linguistics.

\bibitem[{Peng et~al.(2023)Peng, Tian, Yan, Berrebbi, Chang, Li, Shi, Arora, Chen, Sharma, Zhang, Sudo, Shakeel, Jung, Maiti, and Watanabe}]{owsm}
Yifan Peng, Jinchuan Tian, Brian Yan, Dan Berrebbi, Xuankai Chang, Xinjian Li, Jiatong Shi, Siddhant Arora, William Chen, Roshan~S. Sharma, Wangyou Zhang, Yui Sudo, Muhammad Shakeel, Jee{-}weon Jung, Soumi Maiti, and Shinji Watanabe. 2023.
\newblock \href {https://doi.org/10.48550/ARXIV.2309.13876} {Reproducing whisper-style training using an open-source toolkit and publicly available data}.
\newblock \emph{CoRR}, abs/2309.13876.

\bibitem[{Pratap et~al.(2023)Pratap, Tjandra, Shi, Tomasello, Babu, Kundu, Elkahky, Ni, Vyas, Fazel{-}Zarandi, Baevski, Adi, Zhang, Hsu, Conneau, and Auli}]{meta-mms}
Vineel Pratap, Andros Tjandra, Bowen Shi, Paden Tomasello, Arun Babu, Sayani Kundu, Ali Elkahky, Zhaoheng Ni, Apoorv Vyas, Maryam Fazel{-}Zarandi, Alexei Baevski, Yossi Adi, Xiaohui Zhang, Wei{-}Ning Hsu, Alexis Conneau, and Michael Auli. 2023.
\newblock \href {https://doi.org/10.48550/ARXIV.2305.13516} {Scaling speech technology to 1, 000+ languages}.
\newblock \emph{CoRR}, abs/2305.13516.

\bibitem[{Radford et~al.(2023)Radford, Kim, Xu, Brockman, McLeavey, and Sutskever}]{whisper}
Alec Radford, Jong~Wook Kim, Tao Xu, Greg Brockman, Christine McLeavey, and Ilya Sutskever. 2023.
\newblock \href {https://proceedings.mlr.press/v202/radford23a.html} {Robust speech recognition via large-scale weak supervision}.
\newblock In \emph{International Conference on Machine Learning, {ICML} 2023, 23-29 July 2023, Honolulu, Hawaii, {USA}}, volume 202 of \emph{Proceedings of Machine Learning Research}, pages 28492--28518. {PMLR}.

\bibitem[{Rubenstein et~al.(2023)Rubenstein, Asawaroengchai, Nguyen, Bapna, Borsos, de~Chaumont~Quitry, Chen, Badawy, Han, Kharitonov, Muckenhirn, Padfield, Qin, Rozenberg, Sainath, Schalkwyk, Sharifi, Ramanovich, Tagliasacchi, Tudor, Velimirovic, Vincent, Yu, Wang, Zayats, Zeghidour, Zhang, Zhang, Zilka, and Frank}]{audiopalm}
Paul~K. Rubenstein, Chulayuth Asawaroengchai, Duc~Dung Nguyen, Ankur Bapna, Zal{\'{a}}n Borsos, F{\'{e}}lix de~Chaumont~Quitry, Peter Chen, Dalia~El Badawy, Wei Han, Eugene Kharitonov, Hannah Muckenhirn, Dirk Padfield, James Qin, Danny Rozenberg, Tara~N. Sainath, Johan Schalkwyk, Matthew Sharifi, Michelle~Tadmor Ramanovich, Marco Tagliasacchi, Alexandru Tudor, Mihajlo Velimirovic, Damien Vincent, Jiahui Yu, Yongqiang Wang, Vicky Zayats, Neil Zeghidour, Yu~Zhang, Zhishuai Zhang, Lukas Zilka, and Christian~Havn{\o} Frank. 2023.
\newblock \href {https://doi.org/10.48550/arXiv.2306.12925} {Audiopalm: {A} large language model that can speak and listen}.
\newblock \emph{CoRR}, abs/2306.12925.

\bibitem[{Shen et~al.(2023)Shen, Ju, Tan, Liu, Leng, He, Qin, Zhao, and Bian}]{natural-speech-2}
Kai Shen, Zeqian Ju, Xu~Tan, Yanqing Liu, Yichong Leng, Lei He, Tao Qin, Sheng Zhao, and Jiang Bian. 2023.
\newblock \href {https://doi.org/10.48550/arXiv.2304.09116} {Naturalspeech 2: Latent diffusion models are natural and zero-shot speech and singing synthesizers}.
\newblock \emph{CoRR}, abs/2304.09116.

\bibitem[{Steinmetz and Reiss(2021)}]{steinmetz2021pyloudnorm}
Christian~J. Steinmetz and Joshua~D. Reiss. 2021.
\newblock pyloudnorm: {A} simple yet flexible loudness meter in python.
\newblock In \emph{150th AES Convention}.

\bibitem[{Touvron et~al.(2023{\natexlab{a}})Touvron, Lavril, Izacard, Martinet, Lachaux, Lacroix, Rozi{\`{e}}re, Goyal, Hambro, Azhar, Rodriguez, Joulin, Grave, and Lample}]{touvron2023llama}
Hugo Touvron, Thibaut Lavril, Gautier Izacard, Xavier Martinet, Marie{-}Anne Lachaux, Timoth{\'{e}}e Lacroix, Baptiste Rozi{\`{e}}re, Naman Goyal, Eric Hambro, Faisal Azhar, Aur{\'{e}}lien Rodriguez, Armand Joulin, Edouard Grave, and Guillaume Lample. 2023{\natexlab{a}}.
\newblock \href {https://doi.org/10.48550/arXiv.2302.13971} {Llama: Open and efficient foundation language models}.
\newblock \emph{CoRR}, abs/2302.13971.

\bibitem[{Touvron et~al.(2023{\natexlab{b}})Touvron, Martin, Stone, Albert, Almahairi, Babaei, Bashlykov, Batra, Bhargava, Bhosale, Bikel, Blecher, Canton{-}Ferrer, Chen, Cucurull, Esiobu, Fernandes, Fu, Fu, Fuller, Gao, Goswami, Goyal, Hartshorn, Hosseini, Hou, Inan, Kardas, Kerkez, Khabsa, Kloumann, Korenev, Koura, Lachaux, Lavril, Lee, Liskovich, Lu, Mao, Martinet, Mihaylov, Mishra, Molybog, Nie, Poulton, Reizenstein, Rungta, Saladi, Schelten, Silva, Smith, Subramanian, Tan, Tang, Taylor, Williams, Kuan, Xu, Yan, Zarov, Zhang, Fan, Kambadur, Narang, Rodriguez, Stojnic, Edunov, and Scialom}]{llama2}
Hugo Touvron, Louis Martin, Kevin Stone, Peter Albert, Amjad Almahairi, Yasmine Babaei, Nikolay Bashlykov, Soumya Batra, Prajjwal Bhargava, Shruti Bhosale, Dan Bikel, Lukas Blecher, Cristian Canton{-}Ferrer, Moya Chen, Guillem Cucurull, David Esiobu, Jude Fernandes, Jeremy Fu, Wenyin Fu, Brian Fuller, Cynthia Gao, Vedanuj Goswami, Naman Goyal, Anthony Hartshorn, Saghar Hosseini, Rui Hou, Hakan Inan, Marcin Kardas, Viktor Kerkez, Madian Khabsa, Isabel Kloumann, Artem Korenev, Punit~Singh Koura, Marie{-}Anne Lachaux, Thibaut Lavril, Jenya Lee, Diana Liskovich, Yinghai Lu, Yuning Mao, Xavier Martinet, Todor Mihaylov, Pushkar Mishra, Igor Molybog, Yixin Nie, Andrew Poulton, Jeremy Reizenstein, Rashi Rungta, Kalyan Saladi, Alan Schelten, Ruan Silva, Eric~Michael Smith, Ranjan Subramanian, Xiaoqing~Ellen Tan, Binh Tang, Ross Taylor, Adina Williams, Jian~Xiang Kuan, Puxin Xu, Zheng Yan, Iliyan Zarov, Yuchen Zhang, Angela Fan, Melanie Kambadur, Sharan Narang, Aur{\'{e}}lien Rodriguez, Robert Stojnic, Sergey Edunov,
  and Thomas Scialom. 2023{\natexlab{b}}.
\newblock \href {https://doi.org/10.48550/ARXIV.2307.09288} {Llama 2: Open foundation and fine-tuned chat models}.
\newblock \emph{CoRR}, abs/2307.09288.

\bibitem[{Wang et~al.(2021)Wang, Riviere, Lee, Wu, Talnikar, Haziza, Williamson, Pino, and Dupoux}]{wang-etal-2021-voxpopuli}
Changhan Wang, Morgane Riviere, Ann Lee, Anne Wu, Chaitanya Talnikar, Daniel Haziza, Mary Williamson, Juan Pino, and Emmanuel Dupoux. 2021.
\newblock \href {https://doi.org/10.18653/v1/2021.acl-long.80} {{V}ox{P}opuli: A large-scale multilingual speech corpus for representation learning, semi-supervised learning and interpretation}.
\newblock In \emph{Proceedings of the 59th Annual Meeting of the Association for Computational Linguistics and the 11th International Joint Conference on Natural Language Processing (Volume 1: Long Papers)}, pages 993--1003, Online. Association for Computational Linguistics.

\bibitem[{Wang et~al.(2023{\natexlab{a}})Wang, Chen, Wu, Zhang, Zhou, Liu, Chen, Liu, Wang, Li, He, Zhao, and Wei}]{valle}
Chengyi Wang, Sanyuan Chen, Yu~Wu, Ziqiang Zhang, Long Zhou, Shujie Liu, Zhuo Chen, Yanqing Liu, Huaming Wang, Jinyu Li, Lei He, Sheng Zhao, and Furu Wei. 2023{\natexlab{a}}.
\newblock \href {https://doi.org/10.48550/arXiv.2301.02111} {Neural codec language models are zero-shot text to speech synthesizers}.
\newblock \emph{CoRR}, abs/2301.02111.

\bibitem[{Wang et~al.(2023{\natexlab{b}})Wang, Zhou, Zhang, Wu, Liu, Gaur, Chen, Li, and Wei}]{viola}
Tianrui Wang, Long Zhou, Ziqiang Zhang, Yu~Wu, Shujie Liu, Yashesh Gaur, Zhuo Chen, Jinyu Li, and Furu Wei. 2023{\natexlab{b}}.
\newblock \href {https://doi.org/10.48550/arXiv.2305.16107} {Viola: Unified codec language models for speech recognition, synthesis, and translation}.
\newblock \emph{CoRR}, abs/2305.16107.

\bibitem[{Wang et~al.(2023{\natexlab{c}})Wang, Thakker, Chen, Kanda, Eskimez, Chen, Tang, Liu, Li, and Yoshioka}]{speechx}
Xiaofei Wang, Manthan Thakker, Zhuo Chen, Naoyuki Kanda, Sefik~Emre Eskimez, Sanyuan Chen, Min Tang, Shujie Liu, Jinyu Li, and Takuya Yoshioka. 2023{\natexlab{c}}.
\newblock \href {https://doi.org/10.48550/arXiv.2308.06873} {Speechx: Neural codec language model as a versatile speech transformer}.
\newblock \emph{CoRR}, abs/2308.06873.

\bibitem[{Zeghidour et~al.(2022)Zeghidour, Luebs, Omran, Skoglund, and Tagliasacchi}]{soundstream}
Neil Zeghidour, Alejandro Luebs, Ahmed Omran, Jan Skoglund, and Marco Tagliasacchi. 2022.
\newblock \href {https://doi.org/10.1109/TASLP.2021.3129994} {Soundstream: An end-to-end neural audio codec}.
\newblock \emph{{IEEE} {ACM} Trans. Audio Speech Lang. Process.}, 30:495--507.

\bibitem[{Zhang et~al.(2022)Zhang, Roller, Goyal, Artetxe, Chen, Chen, Dewan, Diab, Li, Lin, Mihaylov, Ott, Shleifer, Shuster, Simig, Koura, Sridhar, Wang, and Zettlemoyer}]{zhang2022opt}
Susan Zhang, Stephen Roller, Naman Goyal, Mikel Artetxe, Moya Chen, Shuohui Chen, Christopher Dewan, Mona~T. Diab, Xian Li, Xi~Victoria Lin, Todor Mihaylov, Myle Ott, Sam Shleifer, Kurt Shuster, Daniel Simig, Punit~Singh Koura, Anjali Sridhar, Tianlu Wang, and Luke Zettlemoyer. 2022.
\newblock \href {https://doi.org/10.48550/arXiv.2205.01068} {{OPT:} open pre-trained transformer language models}.
\newblock \emph{CoRR}, abs/2205.01068.

\bibitem[{Zhang et~al.(2023{\natexlab{a}})Zhang, Han, Qin, Wang, Bapna, Chen, Chen, Li, Axelrod, Wang, Meng, Hu, Rosenberg, Prabhavalkar, Park, Haghani, Riesa, Perng, Soltau, Strohman, Ramabhadran, Sainath, Moreno, Chiu, Schalkwyk, Beaufays, and Wu}]{google-usm}
Yu~Zhang, Wei Han, James Qin, Yongqiang Wang, Ankur Bapna, Zhehuai Chen, Nanxin Chen, Bo~Li, Vera Axelrod, Gary Wang, Zhong Meng, Ke~Hu, Andrew Rosenberg, Rohit Prabhavalkar, Daniel~S. Park, Parisa Haghani, Jason Riesa, Ginger Perng, Hagen Soltau, Trevor Strohman, Bhuvana Ramabhadran, Tara~N. Sainath, Pedro~J. Moreno, Chung{-}Cheng Chiu, Johan Schalkwyk, Fran{\c{c}}oise Beaufays, and Yonghui Wu. 2023{\natexlab{a}}.
\newblock \href {https://doi.org/10.48550/ARXIV.2303.01037} {Google {USM:} scaling automatic speech recognition beyond 100 languages}.
\newblock \emph{CoRR}, abs/2303.01037.

\bibitem[{Zhang et~al.(2023{\natexlab{b}})Zhang, Zhou, Wang, Chen, Wu, Liu, Chen, Liu, Wang, Li, He, Zhao, and Wei}]{vallex}
Ziqiang Zhang, Long Zhou, Chengyi Wang, Sanyuan Chen, Yu~Wu, Shujie Liu, Zhuo Chen, Yanqing Liu, Huaming Wang, Jinyu Li, Lei He, Sheng Zhao, and Furu Wei. 2023{\natexlab{b}}.
\newblock \href {https://doi.org/10.48550/arXiv.2303.03926} {Speak foreign languages with your own voice: Cross-lingual neural codec language modeling}.
\newblock \emph{CoRR}, abs/2303.03926.

\end{thebibliography}

\appendix

~\newpage
\section{Literature Review of Speech LMs}
\label{app:literature-review}

The remarkable achievements of large language models (LLMs) in natural language processing (NLP) \citep{DBLP:conf/nips/BrownMRSKDNSSAA20,chowdhery2022palm,zhang2022opt,touvron2023llama,llama2,gpt-4} have served as a powerful impetus for the advancement of foundational models in the realm of speech \citep{whisper,google-usm,meta-mms,seamlessm4t,owsm}, including the emergence and evolution of speech language models \citep{lakhotia-etal-2021-generative, kharitonov-etal-2022-text, audiolm, Speechlmscore, valle, vallex, audiopalm, speechx, voicebox, voxtlm}. 

\autoref{tab:speechlm-papers} compares recent studies about speech LMs from four aspects: speech representation, model architecture, initialization method, and supported tasks. We provide more details in the following sections.

\subsection{Speech representation}
Speech signals can be represented as continuous features or discrete units. Continuous features include spectrograms and neural codec hidden vectors. Discrete units can be further categorized into two types: semantic units and acoustic units. 
Semantic units are derived from self-supervised learning (SSL) or automatic speech recognition (ASR) models through clustering. They are found to mainly capture the linguistic content~\citep{audiolm}, and can thus be used interchangeably with normal text tokens.
Acoustic units are produced by audio codec models through residual vector quantization (RVQ). They capture rich acoustic information like speaker style, emotion, and acoustic environment, making them especially suitable for high-quality speech synthesis. But they are more difficult to model due to the multiple streams from RVQ.

\subsection{Supported tasks}
Conditional speech synthesis is the primary task of speech LMs. As illustrated in \autoref{fig:task}, given semantic units extracted from a content audio, it aims to generate high-quality speech that mimics the style of a short prompt. Our study focuses on this primary task, since other speech generation tasks can be incorporated into this framework.

\subsection{AR vs NAR LMs}

Autoregressive (AR) speech LMs are conditional LMs which predict a sequence of acoustic units given a sequence of semantic units. \autoref{fig:ar-lm} illustrates this inference procedure. Both semantic and acoustic units are derived in an unsupervised manner. Hence, these LMs can be trained on audio-only data without human annotation. Since acoustic units consist of multiple streams, we follow VALL-E~\citep{valle} to predict only the first stream in the AR LM and employ an additional NAR LM to predict the remaining streams. This formulation has been widely used in recent studies~\citep{valle, vallex, viola, polyvoice, speechx}. 
We also consider two variants of AR LMs: one with duplicate semantic units and the other with deduplicated semantic units. The former uses the raw semantic units without extra preprocessing, which leads to a fixed alignment between semantic and acoustic units and thus reduces the length diversity of the synthesized speech. The latter removes consecutive repetitions in semantic units, allowing the AR LM to learn duration information.

Non-autoregressive (NAR) speech LMs predict an entire sequence of continuous features or acoustic units given the corresponding semantic units. \autoref{fig:nar-lm} illustrates the inference procedure. Similar to AR LMs, this formulation does not need human annotation and these models can be trained on audio-only data. 
NaturalSpeech~2~\citep{natural-speech-2} and Voicebox~\citep{voicebox} are two representative NAR LMs. NaturalSpeech 2 learns latent features of a neural codec using a diffusion model, which are converted to waveform with a codec decoder. Voicebox predicts Mel spectrograms using flow matching with the optimal transport path and further synthesizes audios with a HiFi-GAN vocoder~\citep{hifi-gan}. We analyze Voicebox as it shows SOTA performance in a variety of conditional speech synthesis tasks.

\subsection{Analysis of speech LMs}
\label{subsec:analysis-speechlms}

AudioLM~\citep{audiolm} conducts preliminary experiments on its AR LM and 
finds that semantic content and prosodic features are mostly captured by semantic units, while speaker style and recording conditions are from acoustic units.\footnote{It performs quantitative analysis of semantics and speaker style and qualitative analysis of prosody features.}
More recent studies~\citep{soundstorm, make-a-voice, polyvoice} propose various speech LMs in order to achieve zero-shot transfer of vocal styles or speaker emotions.
In this work, we present a systematic investigation of prompt conditioned synthesis based on speech LMs via quantitative analysis, revealing insights into prompt design and unit information which are generalizable to different LM architectures.

\section{Experiments}

\subsection{Speech LM training}
\label{app:lm-training}

We follow TWIST~\citep{twist} to initialize the AR LM with OPT 350M~\citep{zhang2022opt}. Our NAR LM is trained from scratch, which is consistent with Voicebox~\citep{voicebox}. AR LMs are trained using the Adam optimizer~\citep{adam} with a peak learning rate of $0.0002$. They are updated for 200k steps with 20k warmup steps. The NAR LM, Voicebox, is trained for $500$k steps with a learning rate of $0.0001$ and $5$k warmup steps.

\subsection{Effect of heterogeneous prompts}
\label{app:effect-heterogeneous}

\autoref{tab:prompt_het_sim} shows the speaker style similarity between the synthesized audio and each prompt audio. When a single prompt P1 is used, the synthesized audio has a high similarity w.r.t. P1, meaning that the speaker style is well preserved. When a multi-speaker-style prompt (P1+P2) is used, the speaker style similarity decreases drastically, indicating that a heterogeneous prompt hurts speech synthesis. It is interesting to see that the synthesized audio has a higher speaker  style similarity w.r.t. the second prompt segment P2 than the first segment P1, likely because P2 is spatially closer to the generated audio during inference as illustrated in \autoref{fig:ar-lm} and \autoref{fig:nar-lm}. This reveals that speech LMs tend to generate locally coherent audio.

\subsection{Effect of nonstationary prompts}
\label{app:effect-nonstationary}

\autoref{tab:prompt_ns_spkr} shows the speaker style similarity results. When a single prompt P1 is used, the synthesized audio has a high speaker style similarity w.r.t. the prompt, indicating that the speaker style is well preserved. When a multi-style prompt (P1+P2) is used, the speaker style similarity decreases clearly, showing that despite from the same speaker style, multi-emotion prompts adversely affect the preservation of speaker style similarity. 
This indicates that speaker styles and emotions are entangled to some extent. 
The nonstationary nature in prompts distracts speech LMs from capturing speaker style information. We also observe that synthesized audios are more similar to the second prompt segment in terms of speaking style, which is consistent with the previous multi-speaker-style case. This reveals that speech LMs are better at capturing local context than long-range dependencies.

\subsection{Effect of content audio's speaker styles}
\label{app:effect-content-spkr}

When content audios are from F2 who is also female, synthesized audios have the highest similarity w.r.t. the prompt female speaker style F1. When the content speaker style is changed to male speaker styles M1 and M2, synthesized audios demonstrate lower speaker style similarity w.r.t. the same prompt. This reflects that the content audio, represented by semantic units, is also a non-negligible source of speaker style information when speech LMs synthesize audios. This further suggests that semantic units such as HuBERT units carry more acoustic information than expected, which might interfere with style transfer.

\subsection{Analysis of prosody information}
\label{app:analysis-prosody}

We manipulate the acoustic charateristics of prompt or content audios. For example, to study how the prompt's pitch affects speech synthesis, we increase or decrease the pitch of prompt audios, and synthesize a new set of audios with the new prompts. Then, we compute the pitch changes in prompt and generated audios compared to the reference set, and calculate the Pearson correlation between their changes. If the correlation is high, we can infer that the prompt audio is an important source of pitch information. Similarly, we manipulate the speech rate by speeding up or slowing down the prompt/content audios, and manipulate loudness by changing the audio volume with torchaudio\footnote{\url{https://pytorch.org/audio/stable/sox_effects.html}}.

\textbf{More discussions on speech rate.} We find that the unit duration has a strong control over the speech rate. For AR LM with duplicate units and NAR LM, the duration information has been pre-determined and embedded in the sequence of content semantic units. The AR LM with deduplicated units has some flexibility to change the duration, thus mitigating its correlation with the content tempo. 
However, the correlation w.r.t. the prompt tempo is close to zero for all speech LMs, meaning that \textbf{the speech rate cannot be controlled through prompts in current speech LMs.}
We note that none of these models is equipped with explicit duration prediction based on prompt, which is a likely reason of their incapability of capturing prompt's tempo in speech synthesis.

\end{document}